\documentclass[letterpaper, 10 pt, conference]{ieeeconf}  

\IEEEoverridecommandlockouts                              

\usepackage{graphics} 
\usepackage{epsfig} 
\usepackage{times} 
\usepackage{amsmath} 
\usepackage{amssymb}  
\usepackage{booktabs}
\usepackage{array}
\usepackage{tabularx}

\usepackage{makecell}  
\newcolumntype{L}{>{\raggedright\arraybackslash}p{1.75cm}}
\newcolumntype{C}{>{\centering\arraybackslash}X}

\usepackage[table,xcdraw]{xcolor}
\usepackage{soul}
\definecolor{lightviolet}{HTML}{C7A2BA}
\sethlcolor{lightviolet}
\usepackage{subcaption}   

\title{\LARGE \bf
Large Pre-Trained Models for Bimanual Manipulation in 3D
}

\author{Hanna Yurchyk$^{1}$, Wei-Di Chang, Gregory Dudek, David Meger
\thanks{*This work was not supported by any organization}
\thanks{$^{1}$Hanna Yurchyk is with Faculty of Computer Science, McGill University and Mila, Montréal, QC, Canada. Corresponding author: {\tt\small hanna.yurchyk@mail.mcgill.ca}}
}

\begin{document}

\maketitle
\thispagestyle{empty}
\pagestyle{empty}

\begin{abstract}
We investigate the integration of attention maps from a pre-trained Vision Transformer into voxel representations to enhance bimanual robotic manipulation. Specifically, we extract attention maps from DINOv2, a self-supervised ViT model, and interpret them as pixel-level saliency scores over RGB images. These maps are lifted into a 3D voxel grid, resulting in voxel-level semantic cues that are incorporated into a behavior cloning policy. When integrated into a state-of-the-art voxel-based policy, our attention-guided featurization yields an average absolute improvement of 8.2\% and a relative gain of 21.9\% across all tasks in the RLBench bimanual benchmark.

\end{abstract}

\section{INTRODUCTION}\label{sec:1_introduction}

Robotic manipulation remains a core challenge in intelligent robotics \cite{doi:10.1126/science.aat8414, manipulation_ross_tedrake}. Even in well-structured settings, reliable manipulation requires seamless integration of visual perception, geometric reasoning, task planning, and closed-loop control \cite{Diankov2010AutomatedCO}. These challenges are accentuated in bimanual manipulation, where two robot arms must operate in close coordination while sharing the same workspace. Bimanual systems unlock interactions beyond those of single-arm platforms, including simultaneous grasping, handovers, and in-hand re-orientation of a manipulated object \cite{SMITH20121340}. Depending on the system, each arm or a global agent must reason about both the global scene and the fine-grained affordances of the object it handles, while coordinating with the other arm. Importantly, bimanual dexterity is crucial for humanoid and mobile robots with a camera and a dual-arm set-up. Such robots will be expected to perform assistive tasks such as folding laundry \cite{peng2024tiebot, Bersch2011BimanualRC, 5509439}, stocking shelves \cite{Winkler2016KnowledgeEnabledRA}, and providing other assistance in unstructured human environments \cite{4141029, 8187364, Diankov2010AutomatedCO}. Humanoids and mobile platforms with manipulators must reason about the global scene while manipulating small objects precisely with each hand, without self-collision. Thus, they impose even stricter requirements on perception and control than conventional industrial manipulators.

In this paper, we study how semantic information from pre-trained Vision Transformers (ViTs) can enhance voxel-based representations for bimanual manipulation. Building on recent advances in visual reasoning, prior work has shown that features extracted from models like DINO \cite{caron2021dino} and CLIP \cite{radford2021clip} significantly improve downstream performance in tasks such as navigation \cite{jatavallabhula2023conceptfusion} and localization \cite{arnaud2025locate3d}, when lifted from 2D images into 3D representations. Since ViTs capture meaningful semantic cues from raw visual inputs, we hypothesize that embedding these features into structured 3D scene representations will improve performance in robotic manipulation.

This trend has also been explored in imitation learning for manipulation. For instance, Chang et al. \cite{chang2025dvk} showed that DINO-derived keypoints can establish homeomorphic correspondences across object instances, improving generalization in grasping. Di Palo et al. \cite{dipalo2024dinobot} leveraged these keypoints to train a policy from a single demonstration that generalizes effectively to real-world deployment. More broadly, recent works demonstrate that augmenting geometric representations with semantic priors improves the effectiveness of both single and dual-arm policies \cite{gervet2023act3, wilcox2025adapt3r, ze20243ddiffusionpolicygeneralizable, ke20243ddiffuseractorpolicy}. Our work continues on this trend by injecting ViT-derived semantic attention into voxel-based policies, improving task performance without modifying the downstream architecture.

We build on the voxel-based manipulation framework introduced by James et al. \cite{james2022coarsetofineqattention}, who proposed the C2F-ARM policy for single-arm manipulation using coarse-to-fine spatial attention, along with the RLBench benchmark \cite{james2019rlbench}. Shridhar et al. extended this line with PerAct \cite{shridhar2022perceiveractor}, a transformer-based behavioral cloning (BC) agent operating on voxel inputs. More recent methods have further developed voxel policies: Act3D \cite{gervet2023act3} introduced a continuous 3D feature field with coarse-to-fine attention over sampled 3D points, while RVT \cite{goyal2023rvt} proposed a multi-view transformer that fuses re-rendered viewpoints to increase robustness to camera viewpoint variations.

In the bimanual setting, VoxAct-B \cite{liu2024voxactb} and PerAct$^2$ \cite{grotz2024peract2} extend RLBench to support dual-arm manipulation. We adopt VoxAct-B as our baseline since it incorporates several practical improvements such as role identifiers and the use of the Segment Anything Model (SAM) \cite{kirillov2023segment} to crop task-relevant regions.

\begin{figure}[!t]
  \centering
  \includegraphics[width=1\linewidth]{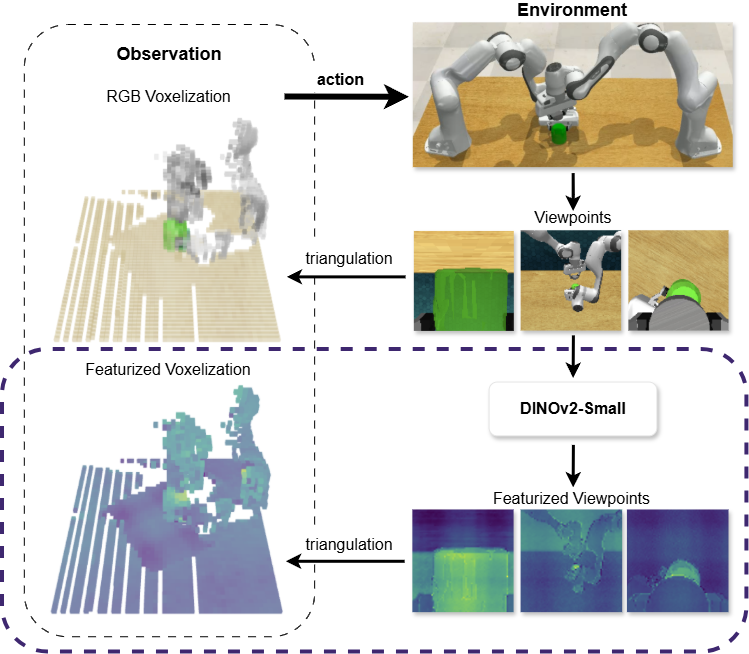}
  \caption{
  Overview of the voxel featurization pipeline. Attention maps are extracted from RGB camera views using DINOv2 \cite{oquab2024dinov2} and projected into a shared 3D voxel grid using multi-view triangulation. RGB and attention features are averaged per voxel.}
  \label{fig:overview}
\end{figure}

These works support a broader insight: combining structured 3D scene representations with high-level semantic priors leads to more robust and generalizable manipulation policies. We propose a lightweight voxel featurization strategy that injects DINOv2 \cite{oquab2024dinov2} derived semantic priors from a single attention head into voxel-based policy learning, generating per-voxel semantic cues for bimanual manipulation. Our main contributions are:
\begin{enumerate}

    \item A pre-processing method that injects ViT-derived attention features into 3D voxel grids with minimal modifications to existing policy frameworks.

    \item Experiments showing that, when combined with VoxAct-B \cite{liu2024voxactb}, our method achieves an \textbf{absolute improvement of 8.2\%} and a \textbf{relative improvement of 21.9\%} across all tasks in the bimanual RLBench.

\end{enumerate}

\section{RELATED WORK}\label{sec:2_related_work}

This section reviews the key research directions that motivate our work. We begin with recent progress in imitation learning (IL) for single-arm manipulation and its extension to dual-arm control. Next, we examine approaches that leverage 3D scene representations to improve spatial reasoning and viewpoint invariance in robotic tasks. Finally, we review recent methods that use large pre-trained visual models to extract semantic features from 2D images and lift them into 3D, enabling richer perception for manipulation.

\subsection{Imitation Learning for Single Arm Control}
Imitation learning (IL) has become a key approach for robotic manipulation, especially when reward functions are sparse. IL systems such as RT-1 \cite{brohan2023rt1} and BC-Z \cite{jang2022bczzeroshottaskgeneralization} show that large-scale datasets of demonstrations can be used to train policies that generalize to a wide range of real-world tasks. However, standard behavior cloning (BC) often suffers from compounding errors and struggles with generalization in complex settings. Recent work addresses these limitations by integrating IL with structured visual inputs and complex supervised learning models to improve policy robustness.

Generative modeling has further advanced this line of research by capturing multimodal action distributions. 3D Diffuser Actor \cite{ke20243ddiffuseractorpolicy} and its bimanual extension Bi3D \cite{ke2024bid} use 3D scene and language inputs to iteratively denoise end-effector trajectories, resulting in generalizable behavior. DP3 \cite{ze20243ddiffusionpolicygeneralizable} shows that compact, point-based 3D encodings can achieve strong performance with relatively few demonstrations. ChainedDiffuser \cite{xian2023chaineddiffuser} combines macro-action planning and diffusion-based trajectory generation into a single model, outperforming both standalone diffusion and keypoint-based methods. 

\subsection{Extending IL to Bimanual Manipulation}

Bimanual manipulation introduces added complexity due to the need for precise coordination between two arms. VoxAct-b \cite{liu2024voxactb} addresses this by using role identifiers (acting and stabilizing arms) and segmentation from SAM to zoom into task-relevant region. PerAct$^2$ \cite{grotz2024peract2} adapts the PerAct transformer to the dual-arm setting by introducing shared network components. Both methods also expand the RLBench benchmark with a broad set of bimanual tasks. Bi3D and RDT-1B \cite{liu2025rdt1bdiffusionfoundationmodel} go further by generating coordinated dual-arm trajectories through diffusion and large-scale pretraining. Outside of end-to-end policies, Gaebert et al. \cite{10769854} learn dual-arm inverse kinematics using conditional generative models, and Chu et al. \cite{10769891} propose a language-based orchestration system for planning long-horizon bimanual tasks.

At a larger scale, generalist frameworks like RT-1 \cite{brohan2023rt1} collect data from many robot platforms to train transformer-based policies with cross-domain transfer capabilities. The $\pi_0$ model \cite{black2024pi0visionlanguageactionflowmodel} integrates a vision-language-action architecture for zero-shot task execution, while Instant Policy \cite{vosylius2025instantpolicyincontextimitation} enables one- or few-shot learning through graph-based diffusion trained on pseudo-demonstrations.

KStar Diffuser builds a spatial-temporal robot graph from the URDF and adds a differentiable kinematics regularizer, showing that integrating these priors during denoising reduces self-collision and infeasible motion near singularities, and reports gains greater than 10\% absolute improvements over strong baselines in simulation and on hardware \cite{lv2025spatialtemporalgraphdiffusionpolicy}. 
 
PPI predicts target gripper \emph{keyposes} and object \emph{pointflow} as interfaces and conditions a diffusion transformer on them to generate continuous dual-arm actions. This balances spatial localization (from keyframes) with flexible motion (from continuous control), achieving an additional $16.1\%$ average success gain on RLBench$^2$ and a $27.5\%$ improvement across four real-world tasks \cite{yang2025gripperkeyposeobjectpointflow}. 

Lift3D and SUGAR indicate that large-scale pretraining can be made 3D-aware without prohibitive cost by reusing 2D backbones and synthetic or simulated supervision. Extending to bimanual settings will require richer multi-arm motion priors, object-to-gripper correspondences, and safety-aware augmentations that respect dual-arm kinematics \cite{jia2024lift3dfoundationpolicylifting,chen2024sugarpretraining3dvisual,yang2025gripperkeyposeobjectpointflow}. 

Across both single and multi-arm settings, a common finding is that combining 3D spatial structure with semantic priors has consistently improved policy robustness

\subsection{3D Representations in Robot Manipulation}

3D representations, such as voxels provide, a structured and spatially aligned format that supports robust manipulation in cluttered scenes. C2F-ARM \cite{james2022coarsetofineqattention} introduces coarse-to-fine Q-attention over voxel grids to guide action selection under sparse rewards. PerAct \cite{shridhar2022perceiveractor} extends this to a transformer-based BC policy that predicts 6-DoF actions directly from voxel inputs. Act3D \cite{gervet2023act3} lifts 2D features into adaptive-resolution 3D point fields, using coarse-to-fine attention to focus on task-relevant areas. RVT \cite{goyal2023rvt} combines multiple camera views through re-rendering and fuses them using a transformer, improving robustness to camera placement and scene variation.

Generative policies also use 3D representations to support conditional action sampling. 3D Diffuser Actor \cite{ke20243ddiffuseractorpolicy}, DP3 \cite{ze20243ddiffusionpolicygeneralizable}, and Bi3D \cite{ke2024bid} use voxel or point-based encodings to generate actions that generalize across object geometries and configurations. Adapt3R \cite{wilcox2025adapt3r} condenses multi-view RGB-D inputs into a point cloud vector representation, which serves as input to an IL model. These approaches demonstrate the value of spatial priors for improving policy performance.

Adapt3R \cite{wilcox2025adapt3r} further improves generalization by integrating 2D semantic features with 3D spatial localization, enabling transfer across viewpoints and robot embodiments. SUGAR shows complementary gains by pretraining on cluttered multi-object point clouds using tasks that align geometry, semantics, and graspability, which leads to better zero-shot recognition and faster fine-tuning for manipulation \cite{chen2024sugarpretraining3dvisual}.

\subsection{ViT Priors for Robotic Manipulation}

Recent advances in large-scale visual pretraining have made it possible to extract high-level semantic information from images and use it to improve robot policies. ViTs like DINO \cite{caron2021dino} and CLIP \cite{radford2021clip} produce pixel or patch-level embeddings that are useful for object recognition, correspondence matching, and alignment. DVK \cite{chang2025dvk} uses DINO features to detect object-centric keypoints that generalize across categories, enabling better generalization in grasping. DINOBot \cite{dipalo2024dinobot} retrieves and aligns demonstrations using ViT features, enabling policy transfer to unseen objects. RSRD \cite{kerr2024robotrobotdoimitating} tracks part-centric object motion in monocular human demonstrations using DINO-based features and 4D reconstruction. These methods show that semantic representations from ViTs can significantly improve learning efficiency, especially in few-shot settings.

On the RLBench single-arm benchmark, Act3D \cite{wilcox2025adapt3r} moves away from voxel representations by using CLIP features to build a 3D point cloud and sample actions from it. Our approach extends this idea to the bimanual RLBench setting by directly embedding DINOv2 \cite{oquab2024dinov2} attention maps into voxel grids, enabling semantic reasoning to guide action selection for dual-arm manipulation.



\begin{figure}[t]
  \centering
  \includegraphics[width=1\linewidth]{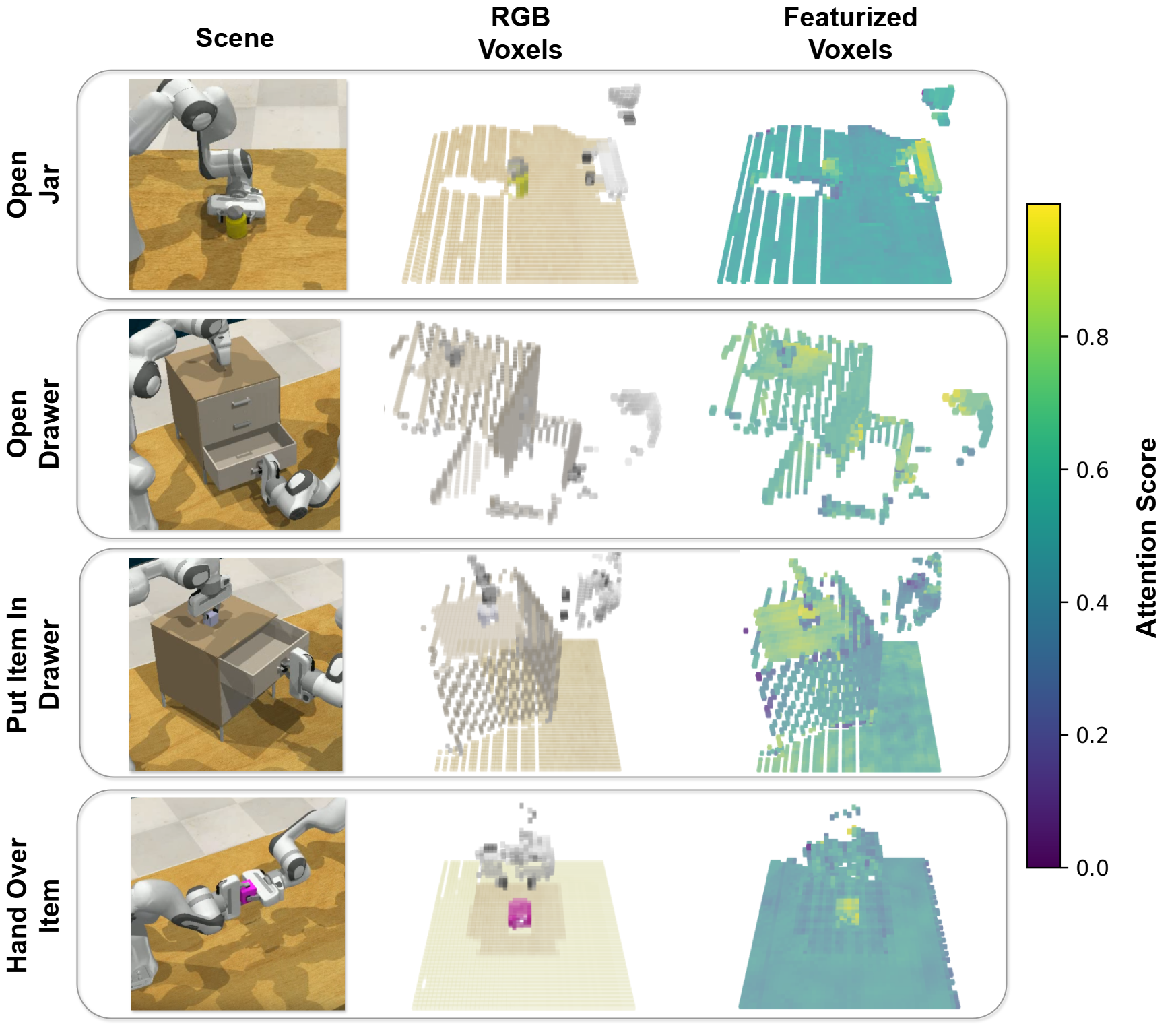}
  \caption{Overview of the bimanual RLBench \cite{liu2024voxactb} manipulation tasks and voxelization examples. Each task includes language instructions specific to the left and right arms, along with variations in the initial position, size, and color of the relevant object. The center column illustrates voxelization from RGB input, while the right column shows the featurized voxel grid augmented with DINOv2 attention. Brighter regions indicate higher attention values and are closer to one in numerical value.}
  \label{fig:all_tasks}
\end{figure}

\section{BACKGROUND}\label{sec:3_background}

\subsection{Benchmark: Bimanual RLBench}

We evaluate our method on the bimanual extension to RLBench \cite{james2019rlbench}, a widely adopted benchmark in the robot learning community \cite{grotz2024peract2, james2019rlbench}. RLBench supports the algorithmic generation of diverse demonstration trajectories and is agnostic to robot morphology, making it compatible with various robotic arms. While originally designed for single-arm manipulation, concurrent extensions by Liu et al. \cite{liu2024voxactb} and Grotz et al. \cite{grotz2024peract2} adapted the benchmark to support bimanual environments.

We build on the bimanual adaptation introduced in VoxAct-B \cite{liu2024voxactb}, which introduces a dual-arm policy architecture with designated acting and stabilizing roles, as well as task-relevant scene cropping via SAM, inspired by prior work in attention-guided cropping \cite{james2022coarsetofineqattention}. Although this version includes only four bimanual tasks (compared to thirteen in \cite{grotz2024peract2}), VoxAct-B provides better performance and more structured policy design.

We adopt the same four tabletop tasks performed with two Franka Emika Panda 7 DoFa arms from VoxAct-B: \textit{open jar}, \textit{open drawer}, \textit{put item in drawer}, and \textit{hand over item} which are illustrated in Figure \ref{fig:all_tasks}. Each task includes variation in object position, size, and color. Language instructions are sampled from a set of templates with paraphrased goals, following the design of \cite{shridhar2022perceiveractor}. For full task details, we refer the reader to \cite{liu2024voxactb}.

\subsection{Baseline Method: VoxAct-B}

Our method is combined with VoxAct-B and the purpose of this section is to provide an overview of the pipeline. For each timestep $t$, the policy receives as input a voxelized scene $\mathbf{v}$, proprioceptive inputs from both arms $\boldsymbol{\rho}$, a language instruction {$\boldsymbol{l}$, and a binary arm identifier $\boldsymbol{\xi} \in \{0, 1\}$ denoting the left or right arm. The voxelized input encodes a $2\times2\times2$ m$^3$ workspace using a $50\times50\times50$ grid. The language instruction is arm-specific, denoted as $\boldsymbol{l} \in \{\boldsymbol{l}_{\text{acting}}, \boldsymbol{l}_{\text{stabilizing}}\}$, and is either retrieved from the dataset during training or selected via a VLM query at evaluation time.

The policy outputs actions for both the acting and stabilizing arms, denoted by $a_t = (a^a_t, a^s_t)$. Each action includes a 3D translation $(x, y, z)$, rotation angles $(\psi, \theta, \phi)$, a binary gripper state $v$, an arm ID $w$, and binary collision indicator $\kappa$. 

Similar to PerAct \cite{shridhar2022perceiveractor}, each policy is implemented as a PerceiverIO \cite{jaegle2022perceiverio}, a transformer-based architecture, followed by MLPs and softmax decoding layers. The predicted Q-values for each output dimension are converted to probabilities using a softmax function, and the highest-scoring action is selected. For each arm, we compute:
\begin{align*}
  &\mathcal{V}_{\mathrm{trans}}
    = \sigma\left(\mathcal{Q}_{\mathrm{trans}}((x,y,z) \mid \cdot)\right)
  && \mathcal{V}_{\mathrm{coll}}
    = \sigma\left(\mathcal{Q}_{\mathrm{coll}}(\kappa \mid \cdot)\right)
  \\
  &\mathcal{V}_{\mathrm{rot}}
    = \sigma\left(\mathcal{Q}_{\mathrm{rot}}((\psi,\theta,\phi) \mid \cdot)\right)
  && \mathcal{V}_{\mathrm{id}}
    = \sigma\left(\mathcal{Q}_{\mathrm{id}}(w \mid \cdot)\right)
  \\
  &\mathcal{V}_{\mathrm{open}}
    = \sigma\left(\mathcal{Q}_{\mathrm{open}}(v \mid \cdot)\right)
  &&
\end{align*}
where $\cdot$ represents the concatenated input $(\mathbf{v}, \boldsymbol{\rho}, \boldsymbol{l}, \boldsymbol{\xi})$ and $\sigma$ is the softmax function. For full implementation details, see \cite{liu2024voxactb}.

During training, demonstration trajectories are split into \textit{keyframes} to reduce sample redundancy and focus on meaningful action states—defined either by a change in gripper state or by termination of end-effector motion \cite{shridhar2022perceiveractor}. These keyframes are stored in a replay buffer and randomly sampled to form training batches.

Each target output is discretized: translation as a one-hot voxel grid $Y_{\text{trans}} \in \mathbb{R}^{H \times W \times D}$, rotation as three one-hot vectors with $R = 5^\circ$ resolution per axis $Y_{\text{rot}} \in \mathbb{R}^{(360/R) \times 3}$, and all binary values (gripper state, collision, arm ID) as 2D one-hot vectors $Y \in \mathbb{R}^2$. A cross-entropy loss is computed for each output channel $i \in \mathcal{X} = \{\text{trans}, \text{rot}, \text{open}, \text{coll}, \text{id}\}$:
\begin{equation}
    \mathcal{L}_i = - \mathbb{E}_{Y_i}[\log \mathcal{V}_i]
\end{equation}
The total loss sums over all output channels and both arms:
\begin{equation}
     \mathcal{L}^{\text{total}} = \sum_{i\in\mathcal{X}} \mathcal{L}_i^{\text{acting}} + \sum_{i\in\mathcal{X}} \mathcal{L}_i^{\text{stabilizing}}
\end{equation}
This supervised loss provides a training signal for voxel-based prediction of dual-arm 6-DoF actions from multimodal inputs.

\begin{figure*}[h!tbp]
  \centering
  \includegraphics[width=\textwidth]{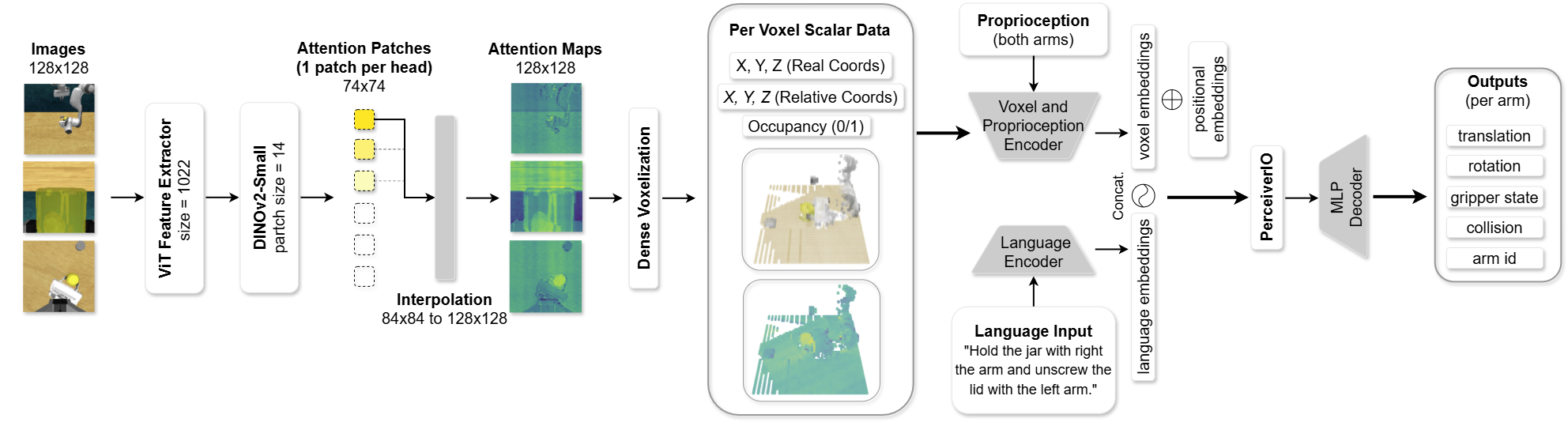}
  \caption{Overview of the voxel featurization and learning pipeline. At each time step \textit{t}, RGB-D observations are processed by DINOv2 to extract attention maps, which are projected into a shared 3D voxel grid alongside RGB features, geometric coordinates, and occupancy values. The voxelized scene is then combined with proprioceptive inputs from both arms. A language instruction is encoded using CLIP and fused with the scene and proprioceptive representations. The resulting input is flattened and passed to a PerceiverIO policy learner, which predicts the action for time step \textit{t+1}.}
  \label{fig:voxelization}
\end{figure*}

\section{METHODOLOGY}\label{sec:4_methodology}

Our method featurizes voxel representations for policy learning with attention maps extracted from a pre-trained ViT. Specifically, we use DINOv2 \cite{oquab2024dinov2}, a self-supervised ViT model trained on large-scale visual data, to produce dense attention maps from RGB inputs. These attention maps are interpreted as maps highlighting important areas of the image and are fused with RGB and depth features into a shared voxel grid representation. This process is shown in Figure \ref{fig:voxelization}.

\paragraph{Visual Feature Extraction}
For each observation, we pass the RGB input through the ViT-S/14 configuration of DINOv2, which generates maps from each of the six self-attention heads of the multi-head attention mechanism from the final transformer layer. Each attention head produces a separate map that highlights distinct spatial patterns in the image. Although these heads are not explicitly grounded in language, they implicitly encode semantic priors such as object boundaries, texture variations, or high-contrast regions.

In our main experiments, we incorporate the first attention head as a lightweight semantic cue, and conduct ablation studies showing that using five attention maps increases the performance by an absolute improvement of 29.0\%. The RGB image is first normalized and then passed through DINOv2. The resulting attention maps are extracted from the final layer as patch-level values in the range $[0, 1]$, where values closer to one represent areas with increased attention. Since these maps are lower-resolution ($74\times74$ pixels), we resize them to the original input resolution ($128\times128$ pixels) using bilinear interpolation. To suppress noise, we apply a soft threshold of 0.6 to the attention maps to remove unwanted artifacts, which typically correspond to high-intensity regions in the feature map \cite{darcet2024visiontransformersneedregisters}.

\paragraph{Voxelization Pipeline}
We extend the voxelization procedure proposed by James et al. \cite{james2019rlbench}, which converts multi-view RGB-D inputs into a dense voxel grid. Given a known camera intrinsics matrix and synchronized depth maps, we ray cast each pixel into a 3D point cloud and associate it with its RGB color and attention score.

To construct the voxel grid, the 3D workspace is defined by bounding coordinates, which is kept as a $2\times2\times2$~m$^3$ cube centered over the workspace, that is discretized into a $50\times50\times50$ grid. Each 3D point from the RGB-D camera is mapped into a corresponding voxel bin based on its spatial coordinates. When multiple points fall into the same voxel, we average their RGB and attention values. The voxel resolution in our setup is approximately 1.2 cm.

Notably, we do not replace RGB features with attention maps to avoid removing color information. Instead, we append the attention score as an additional channel, enhancing the visual representation with supplemental saliency cues.

\paragraph{Voxel Feature Channels}
For each occupied voxel, we compute a fixed set of features capturing both visual and spatial information. These include:
\begin{itemize}
  \item RGB averaged across all points in the voxel (3 channels)
  \item DINOv2 attention score (1 or more channel)
  \item Mean 3D world position $(X, Y, Z)$ of the voxel (3 channels)
  \item Relative grid location $(\bar{X}, \bar{Y}, \bar{Z})$ normalized to $[0, 1]$ (3 channels)
  \item Binary occupancy flag (1 channel),
\end{itemize}
resulting in an 11-dimensional feature vector per voxel. If more attention heads are used, this number scales accordingly (up to 16 features). Empty voxels, or the voxels without corresponding depth points, are zeroed out in all channels except occupancy.

This voxel representation preserves the geometric structure of the scene while enriching it with semantic cues derived from large-scale visual pretraining. Importantly, our featurization requires no fine-tuning of DINOv2 and introduces no architectural changes to the downstream policy. This modularity enables seamless integration into existing voxel-based manipulation pipelines, including the VoxAct-B baseline described in Section \ref{sec:3_background}.

\section{EXPERIMENTS}\label{sec:5_experiments}

We evaluate our method in a learning from demonstration setup on four single-task manipulation scenarios in simulation. All experiments are conducted using the bimanual extension of RLBench.

\subsection{Datasets}

For each task, expert demonstrations are collected using RLBench which uses the Open Motion Planning Library (OMPL) \cite{sucan2012the-open-motion-planning-library} to generate tasks. We generate datasets with 10 and 100 demonstrations per task, accompanied by 25 validation and 25 test episodes with randomized seeds. The camera poses are registered in the simulation environment, allowing consistent multi-view projection. Task variation includes changes in object size, initial pose, color, and associated language instructions \cite{james2019rlbench}.

We follow the data augmentation protocols from \cite{shridhar2022perceiveractor, liu2024voxactb}, applying SE(3) perturbations to enhance generalization. Each demonstration includes proprioceptive state for both arms and RGB-D inputs from three synchronized cameras (front, left wrist, right wrist). Each trajectory spans approximately 100–250 frames. Keyframe extraction is performed after the data collection step.

\subsection{Training}

The full policy (which includes the acting and stabilizing policies) is trained for 1 million steps using a batch size of 1 on two NVIDIA A100 GPUs per task. Training time per task averages 36.5 hours with 100 demonstrations. Checkpoints are saved every 10k steps.

Compared to VoxAct-B, which requires 32 hours of training under identical settings, our method adds negligible overhead while delivering improved performance. In contrast, Bimanual PerActs is reported to require up to 10 days of training per task \cite{liu2024voxactb}, limiting scalability. Due to this significant runtime, this method was not evaluated using 100 demonstrations. PerAct$^2$ \cite{grotz2024peract2} trains in 54 hours per task using 100 demonstrations, while Act3D \cite{gervet2023act3} reports a training time of one day for 200k steps.

\subsection{Baselines}
 
We compare our method against five strong baselines in bimanual RLBenchm, and report their results in Table 4.2:

\paragraph{Diffusion Policy} models visuomotor control as a conditional denoising diffusion process in continuous joint space, enabling stable learning and multimodal action generation from raw visual input \cite{chi2024diffusionpolicyvisuomotorpolicy}. It is designed to operate in both a single and dual arm set-up. 

\paragraph{ACT with Transformers} is a transformer-based imitation learning method that models sequences of joint-space actions to perform fine-grained manipulation from real-world demonstrations \cite{zhao2023learningfinegrainedbimanualmanipulation}. It operates on multi-view RGB-D input and is designed to handle high-precision bimanual tasks with low-cost hardware, but does not incorporate structured 3D scene representations like voxels.

\paragraph{VoxPoser} leverages large language models and vision-language models to compose 3D affordance maps from language instructions, enabling the generation of 6-DoF trajectories without task-specific training \cite{huang2023voxposercomposable3dvalue}. It grounds high-level language reasoning into spatial representations, which are used within a model-based planner for closed-loop manipulation. The original work was for single arm manipulation, and was adapted to a dual-arm setting for benchmarking.

\paragraph{Bimanual PerActs} extends PerAct \cite{shridhar2022perceiveractor} to a two-arm setting with dual-arm input representations and parallel action decoders. Due to the computational cost, it is only evaluated in the 10 demo setting.

\subsection{Evaluation Metric}
In the line of prior work, policies are evaluated based on task success rate, defined as the proportion of episodes in which the agent completes the task as specified by the language instruction. Each success is rated a 100, and each failure 0. 

\subsection{Results}

Each policy is trained using 10 and 100 demonstrations across five random seeds. We select the best performing combination of acting and stabilizing checkpoints based on validation scores over 25 validation episodes and report the mean success rate across 25 unseen test episodes.

As shown in Table~\ref{tab:test-results}, our method consistently outperforms all baselines across tasks and demonstration regimes. Notable \textbf{absolute improvements} over VoxAct-B include:
\begin{itemize}
    \item 15.0\% gain in \textit{open jar} (10 demos)
    \item 14.4\% gain in \textit{open drawer} (100 demos)
    \item 12.0\% gain in \textit{put item in drawer} (10 demos)
    \item 8.8\% improvement on VoxAct-b when trained with 100 demonstrations on \textit{open drawer}, using only 10 demonstrations, highlighting the improved sample efficiency of our approach.
\end{itemize}

\begin{table}[!htbp]
  \centering
  \small                           
  \setlength{\tabcolsep}{2pt}      
  \renewcommand{\arraystretch}{1.3} 

  \begin{tabularx}{\columnwidth}{@{} L *{8}{C} @{}}
    \toprule
    \textbf{Method}
      & \multicolumn{2}{c}{\makecell{\textbf{Open}\\\textbf{Jar}}}
      & \multicolumn{2}{c}{\makecell{\textbf{Open}\\\textbf{Drawer}}}
      & \multicolumn{2}{c}{\makecell{\textbf{Put Item}\\\textbf{in Drawer}}}
      & \multicolumn{2}{c}{\makecell{\textbf{Hand Over}\\\textbf{Item}}} \\
    Number of Demos 
      & 10 & 100 & 10 & 100 & 10 & 100 & 10 & 100 \\
    \midrule
    Diffusion Policy        & 4.8  & 21.6 & 4.8  & 5.6  & 2.4  & 4.8  & 0.0 & 0.0 \\
    ACT with Transformers   & 4.0  & 30.4 & 12.8 & 28.0 & 8.8  & 44.8 & 1.6 & 7.2 \\
    VoxPoser                & 8.0  & 8.0  & 32.0 & 32.0 & 4.0  & 4.0  & 0.0 & 0.0 \\
    Bimanual PerActs        & 8.0  & --   & 36.8 & --   & 5.6  & --   & 0.0 & --  \\
    VoxAct-B                & 40.0 & 59.2 & 73.6 & 72.8 & 39.2 & 49.6 & 19.2& 14.4\\
    \textbf{Ours}   & \textbf{55.8} & \textbf{60.4} & \textbf{81.6} & \textbf{87.2} & \textbf{51.2} & \textbf{55.2} & \textbf{20.8} & \textbf{22.4} \\
    \bottomrule
  \end{tabularx}
  
  \caption{Success rates (\%) across various tasks for single-task policies when trained with 10 and 100 demonstrations.}
  
  \label{tab:test-results}
\end{table}

Performance on \textit{hand over item} remains low for all methods. In PerAct$^2$, related tasks such as \textit{handover} and \textit{handover easy} achieve 11.0\% and 41.0\% success respectively \cite{grotz2024peract2}. Bi3D reports 19.0\% and 20.0\% on the same tasks \cite{ke2024bid}. Our results fall within this performance range, suggesting that existing architectures struggle with precise coordination required for handoff behavior, and motivating further work on improving representations or better arm communication.

\subsection{Ablations}

We evaluate the effect of using different numbers of attention heads from DINOv2 as input features. As shown in Table \ref{tab:ablations}, including just one attention map improves performance over the RGB-only baseline by 15\%. With three and five heads, performance increases further, reaching a 69.8\% success rate on \textit{open jar}. \textbf{This represents a 29.0\% absolute gain over the baseline task.}

Due to resource constraints, we limit our ablation to a subset of attention heads, but future work may explore their combinatorial effects and semantic diversity more thoroughly.

\begin{table}[!htbp]
  \centering 
  \begin{tabular}{c c}
    \toprule
    \makecell{\textbf{Number of} \\ \textbf{Attention Maps}} 
      & \makecell{\textbf{Open Jar} \\ \textbf{Success Rates}} \\
    \midrule
    0 & 40.8 \\
    1 & 55.8 \\
    3 & 64.4 \\
    5 & \textbf{69.8} \\
    \bottomrule
  \end{tabular}
  \caption{Ablations showing success rates (\%) when using multiple attention maps extracted from DINOv2 on \textit{open jar} task.}%
  \label{tab:ablations}
\end{table}

\subsection{Training Loss Curves}

Figure \ref{fig:avg_train_loss} shows the total training loss averaged over five seeds for all tasks. The consistent downward trend confirms that adding attention maps does not destabilize training. Loss curves are smoothed for readability.

\begin{figure}[!htbp]
  \centering
  \includegraphics[width=1\linewidth]{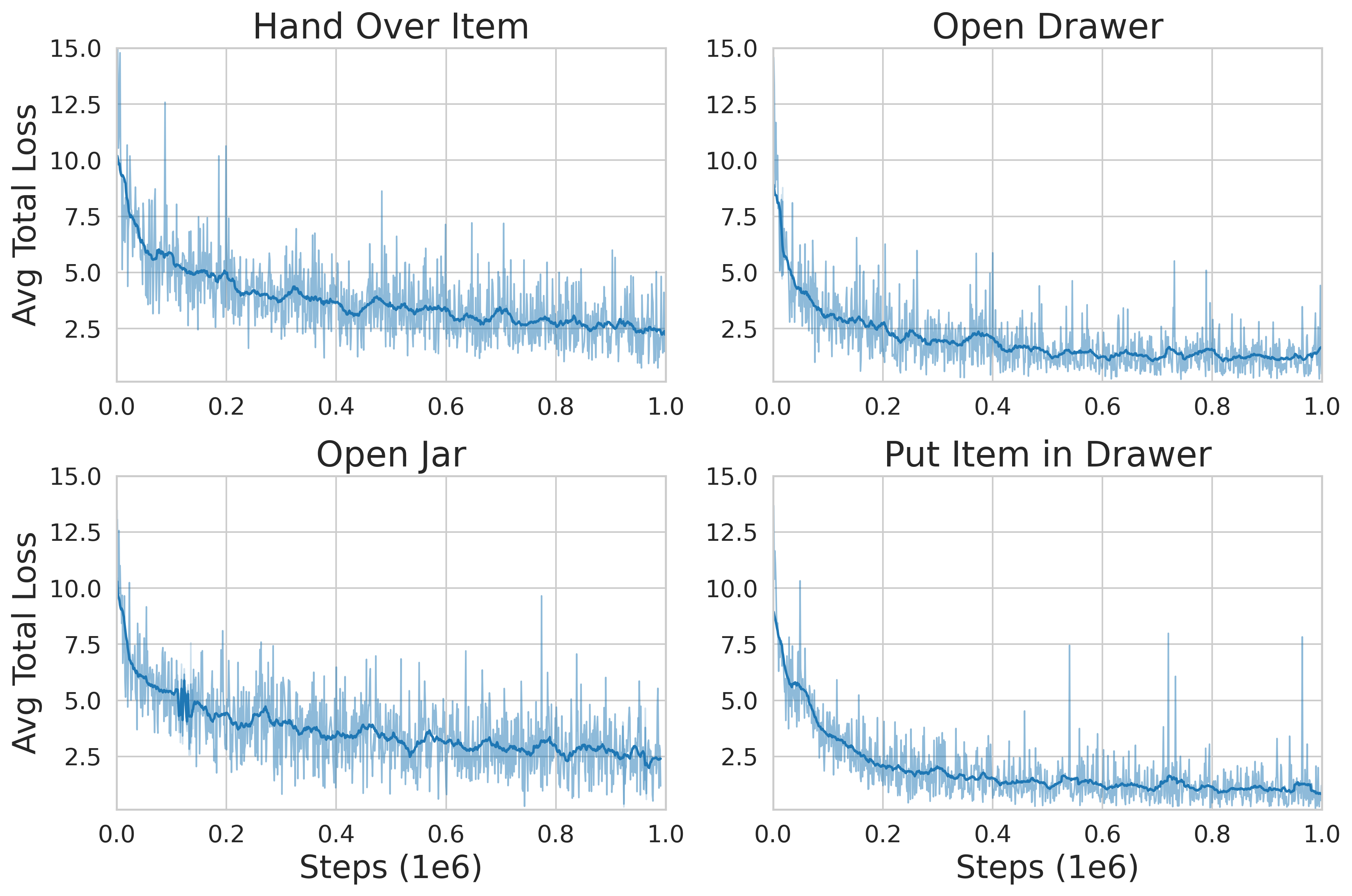}
  \caption{Training loss averaged over five seeds for all tasks, trained on 10 demonstrations. The curves are smoothed for better interpretation.}
  \label{fig:avg_train_loss}
\end{figure}

\subsection{Limitations and Future Work}

While our method achieves notable improvements in task success, several limitations remain. First, the DINOv2 attention heads are not explicitly interpretable, and their contribution to performance varies. Although our ablation results indicate that multiple heads improve learning, understanding which heads encode task-relevant semantics and why remains an open question. Future work may explore attention selection strategies or adapt ViT representations to better suit robotic manipulation.

Second, dexterous bimanual coordination remains a key challenge. Tasks such as \textit{hand over item}, which require precise timing and spatial alignment between arms, continue to show low success rates across methods. This suggests that the current voxel resolution and action modeling may be insufficient for tasks involving close physical interaction, and highlights the need for finer representations or hierarchical dual-arm planning frameworks.

Third, the computational cost of training and evaluation remains high. Like prior work \cite{shridhar2022perceiveractor, liu2024voxactb}, our results are limited to a subset of training seeds and tasks due to hardware constraints. More comprehensive evaluation protocols, including task robustness and failure recovery, are important directions for future benchmarking.

Finally, although ViT-derived features have shown promise in real-world setups \cite{dipalo2024dinobot}, our method is currently validated only in simulation. Bridging this gap will require addressing domain shift and sensor noise, as well as adapting the voxelization and feature extraction pipeline to real camera inputs.

\section{CONCLUSION}\label{sec:6_conclusion}

We introduced a method for enhancing voxel-based policy learning in bimanual robot manipulation by integrating semantic priors from pre-trained visual transformers. By injecting attention maps from DINOv2 into the voxel representation, our approach enriches 3D scene understanding without modifying the downstream policy architecture. Evaluated on the RLBench bimanual benchmark, our method achieves consistent improvements across multiple tasks, with notable gains in both 10 and 100 demonstration training cases. 

Ablation studies confirm that incorporating multiple attention maps leads to substantial gains, highlighting the utility of visual transformer features in 3D spatial reasoning. Our results show that pre-trained ViTs can meaningfully contribute to action learning when embedded into 3D control pipelines.

This work represents a step towards unifying semantic perception and structured spatial representations in robot learning, with the goal of enabling more capable and generalizable manipulation methods.





\bibliography{references}

\begin{thebibliography}{10}
\providecommand{\url}[1]{#1}
\csname url@samestyle\endcsname
\providecommand{\newblock}{\relax}
\providecommand{\bibinfo}[2]{#2}
\providecommand{\BIBentrySTDinterwordspacing}{\spaceskip=0pt\relax}
\providecommand{\BIBentryALTinterwordstretchfactor}{4}
\providecommand{\BIBentryALTinterwordspacing}{\spaceskip=\fontdimen2\font plus
\BIBentryALTinterwordstretchfactor\fontdimen3\font minus \fontdimen4\font\relax}
\providecommand{\BIBforeignlanguage}[2]{{%
\expandafter\ifx\csname l@#1\endcsname\relax
\typeout{** WARNING: IEEEtran.bst: No hyphenation pattern has been}%
\typeout{** loaded for the language `#1'. Using the pattern for}%
\typeout{** the default language instead.}%
\else
\language=\csname l@#1\endcsname
\fi
#2}}
\providecommand{\BIBdecl}{\relax}
\BIBdecl

\bibitem{doi:10.1126/science.aat8414}
\BIBentryALTinterwordspacing
A.~Billard and D.~Kragic, ``Trends and challenges in robot manipulation,'' \emph{Science}, vol. 364, no. 6446, p. eaat8414, 2019. [Online]. Available: \url{https://www.science.org/doi/abs/10.1126/science.aat8414}
\BIBentrySTDinterwordspacing

\bibitem{manipulation_ross_tedrake}
\BIBentryALTinterwordspacing
R.~Tedrake, \emph{Robotic Manipulation}.\hskip 1em plus 0.5em minus 0.4em\relax Course Notes for MIT 6.421, 2024. [Online]. Available: \url{http://manipulation.mit.edu}
\BIBentrySTDinterwordspacing

\bibitem{Diankov2010AutomatedCO}
\BIBentryALTinterwordspacing
R.~Diankov, ``Automated construction of robotic manipulation programs,'' Ph.D. dissertation, Carnegie Mellon University, The Robotics Institute, 2010. [Online]. Available: \url{https://www.ri.cmu.edu/publications/automated-construction-of-robotic-manipulation-programs/}
\BIBentrySTDinterwordspacing

\bibitem{SMITH20121340}
\BIBentryALTinterwordspacing
C.~Smith, Y.~Karayiannidis, L.~Nalpantidis, X.~Gratal, P.~Qi, D.~V. Dimarogonas, and D.~Kragic, ``Dual arm manipulation—a survey,'' \emph{Robotics and Autonomous Systems}, vol.~60, no.~10, pp. 1340--1353, 2012. [Online]. Available: \url{https://www.sciencedirect.com/science/article/pii/S092188901200108X}
\BIBentrySTDinterwordspacing

\bibitem{peng2024tiebot}
\BIBentryALTinterwordspacing
W.~Peng, J.~Lv, Y.~Zeng, H.~Chen, S.~Zhao, J.~Sun, C.~Lu, and L.~Shao, ``Tiebot: Learning to knot a tie from visual demonstration through a real-to-sim-to-real approach,'' in \emph{Conference on Robot Learning, 6-9 November 2024, Munich, Germany}, ser. Proceedings of Machine Learning Research, P.~Agrawal, O.~Kroemer, and W.~Burgard, Eds., vol. 270.\hskip 1em plus 0.5em minus 0.4em\relax {PMLR}, 2024, pp. 318--339. [Online]. Available: \url{https://proceedings.mlr.press/v270/peng25a.html}
\BIBentrySTDinterwordspacing

\bibitem{Bersch2011BimanualRC}
C.~Bersch, B.~Pitzer, and S.~Kammel, ``Bimanual robotic cloth manipulation for laundry folding,'' in \emph{2011 IEEE/RSJ International Conference on Intelligent Robots and Systems}, 2011, pp. 1413--1419.

\bibitem{5509439}
J.~Maitin-Shepard, M.~Cusumano-Towner, J.~Lei, and P.~Abbeel, ``Cloth grasp point detection based on multiple-view geometric cues with application to robotic towel folding,'' in \emph{2010 IEEE International Conference on Robotics and Automation}, 2010, pp. 2308--2315.

\bibitem{Winkler2016KnowledgeEnabledRA}
\BIBentryALTinterwordspacing
J.~O. Winkler, F.~B{\'a}lint-Bencz{\'e}di, T.~Wiedemeyer, M.~Beetz, N.~Vaskevicius, C.~A. Mueller, T.~Fromm, and A.~Birk, ``Knowledge-enabled robotic agents for shelf replenishment in cluttered retail environments: (extended abstract),'' in \emph{Adaptive Agents and Multi-Agent Systems}, 2016. [Online]. Available: \url{https://api.semanticscholar.org/CorpusID:16247186}
\BIBentrySTDinterwordspacing

\bibitem{4141029}
C.~C. Kemp, A.~Edsinger, and E.~Torres-Jara, ``Challenges for robot manipulation in human environments [grand challenges of robotics],'' \emph{IEEE Robotics \& Automation Magazine}, vol.~14, no.~1, pp. 20--29, 2007.

\bibitem{8187364}
M.~A. Goodrich and A.~C. Schultz, ``Human-robot interaction: A survey,'' in \emph{Human-Robot Interaction: A Survey}, 2008.

\bibitem{caron2021dino}
\BIBentryALTinterwordspacing
M.~Caron, H.~Touvron, I.~Misra, H.~J\'egou, J.~Mairal, P.~Bojanowski, and A.~Joulin, ``Emerging properties in self-supervised vision transformers,'' in \emph{Proceedings of the IEEE/CVF International Conference on Computer Vision (ICCV)}, 2021, pp. 9650--9660. [Online]. Available: \url{https://arxiv.org/abs/2104.14294}
\BIBentrySTDinterwordspacing

\bibitem{radford2021clip}
\BIBentryALTinterwordspacing
A.~Radford, J.~W. Kim, C.~Hallacy, A.~Ramesh, G.~Goh, S.~Agarwal, G.~Sastry, A.~Askell, P.~Mishkin, J.~Clark, G.~Krueger, and I.~Sutskever, ``Learning transferable visual models from natural language supervision,'' in \emph{Proceedings of the 38th International Conference on Machine Learning, {ICML} 2021, 18-24 July 2021, Virtual Event}, ser. Proceedings of Machine Learning Research, M.~Meila and T.~Zhang, Eds., vol. 139.\hskip 1em plus 0.5em minus 0.4em\relax {PMLR}, 2021, pp. 8748--8763. [Online]. Available: \url{http://proceedings.mlr.press/v139/radford21a.html}
\BIBentrySTDinterwordspacing

\bibitem{jatavallabhula2023conceptfusion}
\BIBentryALTinterwordspacing
K.~Jatavallabhula, A.~Kuwajerwala, Q.~Gu, M.~Omama, T.~Chen, S.~Li, G.~Iyer, S.~Saryazdi, N.~Keetha, A.~Tewari, J.~Tenenbaum, C.~{de Melo}, M.~Krishna, L.~Paull, F.~Shkurti, and A.~Torralba, ``Conceptfusion: Open-set multimodal 3d mapping,'' \emph{Robotics: Science and Systems (RSS)}, 2023. [Online]. Available: \url{https://arxiv.org/abs/2302.07241}
\BIBentrySTDinterwordspacing

\bibitem{arnaud2025locate3d}
\BIBentryALTinterwordspacing
P.~McVay, S.~Arnaud, A.~Martin, A.~Majumdar, K.~M. Jatavallabhula, P.~Thomas, R.~Partsey, D.~Dugas, A.~Gejji, A.~Sax, V.-P. Berges, M.~Henaff, A.~Jain, A.~Cao, I.~Prasad, M.~Kalakrishnan, M.~Rabbat, N.~Ballas, M.~Assran, O.~Maksymets, A.~Rajeswaran, and F.~Meier, ``{LOCATE} 3d: Real-world object localization via self-supervised learning in 3d,'' in \emph{Forty-second International Conference on Machine Learning}, 2025. [Online]. Available: \url{https://ai.meta.com/research/publications/locate-3d-real-world-object-localization-via-self-supervised-learning-in-3d}
\BIBentrySTDinterwordspacing

\bibitem{chang2025dvk}
\BIBentryALTinterwordspacing
W.-D. Chang, F.~Hogan, S.~Fujimoto, D.~Meger, and G.~Dudek, ``Generalizable imitation learning through pre-trained representations,'' in \emph{IEEE International Conference on Robotics and Automation (ICRA)}, 2025. [Online]. Available: \url{https://arxiv.org/abs/2311.09350}
\BIBentrySTDinterwordspacing

\bibitem{dipalo2024dinobot}
\BIBentryALTinterwordspacing
N.~D. Palo and E.~Johns, ``Dinobot: Robot manipulation via retrieval and alignment with vision foundation models,'' in \emph{{IEEE} International Conference on Robotics and Automation, {ICRA} 2024, Yokohama, Japan, May 13-17, 2024}.\hskip 1em plus 0.5em minus 0.4em\relax {IEEE}, 2024, pp. 2798--2805. [Online]. Available: \url{https://doi.org/10.1109/ICRA57147.2024.10610923}
\BIBentrySTDinterwordspacing

\bibitem{gervet2023act3}
\BIBentryALTinterwordspacing
T.~Gervet, Z.~Xian, N.~Gkanatsios, and K.~Fragkiadaki, ``Act3d: 3d feature field transformers for multi-task robotic manipulation,'' in \emph{Proceedings of The 7th Conference on Robot Learning}, ser. Proceedings of Machine Learning Research, J.~Tan, M.~Toussaint, and K.~Darvish, Eds., vol. 229.\hskip 1em plus 0.5em minus 0.4em\relax PMLR, 2023, pp. 3949--3965. [Online]. Available: \url{https://proceedings.mlr.press/v229/gervet23a.html}
\BIBentrySTDinterwordspacing

\bibitem{wilcox2025adapt3r}
\BIBentryALTinterwordspacing
A.~Wilcox, M.~Ghanem, M.~Moghani, P.~Barroso, B.~Joffe, and A.~Garg, ``Adapt3r: Adaptive 3d scene representation for domain transfer in imitation learning,'' \emph{CoRR}, vol. abs/2503.04877, 2025. [Online]. Available: \url{https://doi.org/10.48550/arXiv.2503.04877}
\BIBentrySTDinterwordspacing

\bibitem{ze20243ddiffusionpolicygeneralizable}
Y.~Ze, G.~Zhang, K.~Zhang, C.~Hu, M.~Wang, and H.~Xu, ``{3D Diffusion Policy: Generalizable Visuomotor Policy Learning via Simple 3D Representations},'' in \emph{Proceedings of Robotics: Science and Systems}, Delft, Netherlands, 2024.

\bibitem{ke20243ddiffuseractorpolicy}
\BIBentryALTinterwordspacing
T.-W. Ke, N.~Gkanatsios, and K.~Fragkiadaki, ``3d diffuser actor: Policy diffusion with 3d scene representations,'' in \emph{Proceedings of The 8th Conference on Robot Learning}, ser. Proceedings of Machine Learning Research, P.~Agrawal, O.~Kroemer, and W.~Burgard, Eds., vol. 270.\hskip 1em plus 0.5em minus 0.4em\relax PMLR, 2025, pp. 1949--1974. [Online]. Available: \url{https://proceedings.mlr.press/v270/ke25a.html}
\BIBentrySTDinterwordspacing

\bibitem{james2022coarsetofineqattention}
S.~James, K.~Wada, T.~Laidlow, and A.~J. Davison, ``Coarse-to-fine q-attention: Efficient learning for visual robotic manipulation via discretisation,'' in \emph{2022 IEEE/CVF Conference on Computer Vision and Pattern Recognition (CVPR)}, 2022, pp. 13\,729--13\,738.

\bibitem{james2019rlbench}
S.~James, Z.~Ma, D.~R. Arrojo, and A.~J. Davison, ``Rlbench: The robot learning benchmark \& learning environment,'' \emph{IEEE Robotics and Automation Letters}, vol.~5, no.~2, pp. 3019--3026, 2020.

\bibitem{shridhar2022perceiveractor}
\BIBentryALTinterwordspacing
M.~Shridhar, L.~Manuelli, and D.~Fox, ``Perceiver-actor: A multi-task transformer for robotic manipulation,'' in \emph{Proceedings of The 6th Conference on Robot Learning}, ser. Proceedings of Machine Learning Research, K.~Liu, D.~Kulic, and J.~Ichnowski, Eds., vol. 205.\hskip 1em plus 0.5em minus 0.4em\relax PMLR, 2023, pp. 785--799. [Online]. Available: \url{https://proceedings.mlr.press/v205/shridhar23a.html}
\BIBentrySTDinterwordspacing

\bibitem{goyal2023rvt}
\BIBentryALTinterwordspacing
A.~Goyal, J.~Xu, Y.~Guo, V.~Blukis, Y.-W. Chao, and D.~Fox, ``Rvt: Robotic view transformer for 3d object manipulation,'' in \emph{Proceedings of The 7th Conference on Robot Learning}, ser. Proceedings of Machine Learning Research, J.~Tan, M.~Toussaint, and K.~Darvish, Eds., vol. 229.\hskip 1em plus 0.5em minus 0.4em\relax PMLR, 2023, pp. 694--710. [Online]. Available: \url{https://proceedings.mlr.press/v229/goyal23a.html}
\BIBentrySTDinterwordspacing

\bibitem{liu2024voxactb}
I.-C.~A. Liu, S.~He, D.~Seita, and G.~S. Sukhatme, ``Voxact-b: Voxel-based acting and stabilizing policy for bimanual manipulation,'' in \emph{Proceedings of The 8th Conference on Robot Learning}, ser. Proceedings of Machine Learning Research, P.~Agrawal, O.~Kroemer, and W.~Burgard, Eds., vol. 270.\hskip 1em plus 0.5em minus 0.4em\relax PMLR, 2025, pp. 4354--4370.

\bibitem{grotz2024peract2}
\BIBentryALTinterwordspacing
M.~Grotz, M.~Shridhar, Y.-W. Chao, T.~Asfour, and D.~Fox, ``Peract2: Benchmarking and learning for robotic bimanual manipulation tasks,'' in \emph{CoRL 2024 Workshop on Whole-body Control and Bimanual Manipulation: Applications in Humanoids and Beyond}, 2024. [Online]. Available: \url{https://arxiv.org/abs/2407.00278}
\BIBentrySTDinterwordspacing

\bibitem{kirillov2023segment}
A.~Kirillov, E.~Mintun, N.~Ravi, H.~Mao, C.~Rolland, L.~Gustafson, T.~Xiao, S.~Whitehead, A.~C. Berg, W.-Y. Lo, P.~Dollar, and R.~Girshick, ``Segment anything,'' in \emph{Proceedings of the IEEE/CVF International Conference on Computer Vision (ICCV)}, 2023, pp. 4015--4026.

\bibitem{oquab2024dinov2}
\BIBentryALTinterwordspacing
M.~Oquab, T.~Darcet, T.~Moutakanni, H.~V. Vo, M.~Szafraniec, V.~Khalidov, P.~Fernandez, D.~HAZIZA, F.~Massa, A.~El-Nouby, M.~Assran, N.~Ballas, W.~Galuba, R.~Howes, P.-Y. Huang, S.-W. Li, I.~Misra, M.~Rabbat, V.~Sharma, G.~Synnaeve, H.~Xu, H.~Jegou, J.~Mairal, P.~Labatut, A.~Joulin, and P.~Bojanowski, ``{DINO}v2: Learning robust visual features without supervision,'' \emph{Transactions on Machine Learning Research}, 2024, featured Certification. [Online]. Available: \url{https://openreview.net/forum?id=a68SUt6zFt}
\BIBentrySTDinterwordspacing

\bibitem{brohan2023rt1}
\BIBentryALTinterwordspacing
A.~Brohan, N.~Brown, J.~Carbajal, Y.~Chebotar, J.~Dabis, C.~Finn, K.~Gopalakrishnan, K.~Hausman, A.~Herzog, J.~Hsu, J.~Ibarz, B.~Ichter, A.~Irpan, T.~Jackson, S.~Jesmonth, N.~J. Joshi, R.~Julian, D.~Kalashnikov, Y.~Kuang, I.~Leal, K.~Lee, S.~Levine, Y.~Lu, U.~Malla, D.~Manjunath, I.~Mordatch, O.~Nachum, C.~Parada, J.~Peralta, E.~Perez, K.~Pertsch, J.~Quiambao, K.~Rao, M.~S. Ryoo, G.~Salazar, P.~Sanketi, K.~Sayed, J.~Singh, S.~Sontakke, A.~Stone, C.~Tan, H.~T. Tran, V.~Vanhoucke, S.~Vega, Q.~Vuong, F.~Xia, T.~Xiao, P.~Xu, S.~Xu, T.~Yu, and B.~Zitkovich, ``{RT-1:} robotics transformer for real-world control at scale,'' \emph{CoRR}, vol. abs/2212.06817, 2022. [Online]. Available: \url{https://doi.org/10.48550/arXiv.2212.06817}
\BIBentrySTDinterwordspacing

\bibitem{jang2022bczzeroshottaskgeneralization}
\BIBentryALTinterwordspacing
E.~Jang, A.~Irpan, M.~Khansari, D.~Kappler, F.~Ebert, C.~Lynch, S.~Levine, and C.~Finn, ``Bc-z: Zero-shot task generalization with robotic imitation learning,'' in \emph{Proceedings of the 5th Conference on Robot Learning}, ser. Proceedings of Machine Learning Research, A.~Faust, D.~Hsu, and G.~Neumann, Eds., vol. 164.\hskip 1em plus 0.5em minus 0.4em\relax PMLR, 2022, pp. 991--1002. [Online]. Available: \url{https://proceedings.mlr.press/v164/jang22a.html}
\BIBentrySTDinterwordspacing

\bibitem{ke2024bid}
\BIBentryALTinterwordspacing
T.-W. Ke, N.~Gkanatsios, J.~Xu, and K.~Fragkiadaki, ``Bi3d diffuser actor: 3d policy diffusion for bi-manual robot manipulation,'' in \emph{CoRL 2024 Workshop on Mastering Robot Manipulation in a World of Abundant Data}, 2024. [Online]. Available: \url{https://openreview.net/forum?id=xcBz0l6yfa}
\BIBentrySTDinterwordspacing

\bibitem{xian2023chaineddiffuser}
\BIBentryALTinterwordspacing
Z.~Xian, N.~Gkanatsios, T.~Gervet, T.-W. Ke, and K.~Fragkiadaki, ``Chaineddiffuser: Unifying trajectory diffusion and keypose prediction for robotic manipulation,'' in \emph{7th Annual Conference on Robot Learning}, 2023. [Online]. Available: \url{https://openreview.net/forum?id=W0zgY2mBTA8}
\BIBentrySTDinterwordspacing

\bibitem{liu2025rdt1bdiffusionfoundationmodel}
\BIBentryALTinterwordspacing
S.~Liu, L.~Wu, B.~Li, H.~Tan, H.~Chen, Z.~Wang, K.~Xu, H.~Su, and J.~Zhu, ``{RDT-1B:} a diffusion foundation model for bimanual manipulation,'' in \emph{The Thirteenth International Conference on Learning Representations, {ICLR} 2025, Singapore, April 24-28, 2025}.\hskip 1em plus 0.5em minus 0.4em\relax OpenReview.net, 2025. [Online]. Available: \url{https://openreview.net/forum?id=yAzN4tz7oI}
\BIBentrySTDinterwordspacing

\bibitem{10769854}
C.~Gaebert and U.~Thomas, ``Generating dual-arm inverse kinematics solutions using latent variable models,'' in \emph{2024 IEEE-RAS 23rd International Conference on Humanoid Robots (Humanoids)}, 2024.

\bibitem{10769891}
K.~Chu, X.~Zhao, C.~Weber, M.~Li, W.~Lu, and S.~Wermter, ``Large language models for orchestrating bimanual robots,'' in \emph{2024 IEEE-RAS 23rd International Conference on Humanoid Robots (Humanoids)}, 2024.

\bibitem{black2024pi0visionlanguageactionflowmodel}
\BIBentryALTinterwordspacing
K.~Black, N.~Brown, D.~Driess, A.~Esmail, M.~Equi, C.~Finn, N.~Fusai, L.~Groom, K.~Hausman, B.~Ichter, S.~Jakubczak, T.~Jones, L.~Ke, S.~Levine, A.~Li-Bell, M.~Mothukuri, S.~Nair, K.~Pertsch, L.~X. Shi, J.~Tanner, Q.~Vuong, A.~Walling, H.~Wang, and U.~Zhilinsky, ``$\pi_0$: A vision-language-action flow model for general robot control,'' 2024. [Online]. Available: \url{https://arxiv.org/abs/2410.24164}
\BIBentrySTDinterwordspacing

\bibitem{vosylius2025instantpolicyincontextimitation}
\BIBentryALTinterwordspacing
V.~Vosylius and E.~Johns, ``Instant policy: In-context imitation learning via graph diffusion,'' in \emph{The Thirteenth International Conference on Learning Representations, {ICLR} 2025, Singapore, April 24-28, 2025}.\hskip 1em plus 0.5em minus 0.4em\relax OpenReview.net, 2025. [Online]. Available: \url{https://openreview.net/forum?id=je3GZissZc}
\BIBentrySTDinterwordspacing

\bibitem{lv2025spatialtemporalgraphdiffusionpolicy}
Q.~Lv, H.~Li, X.~Deng, R.~Shao, Y.~Li, J.~Hao, L.~Gao, M.~Y. Wang, and L.~Nie, ``Spatial-temporal graph diffusion policy with kinematic modeling for bimanual robotic manipulation,'' in \emph{{IEEE/CVF} Conference on Computer Vision and Pattern Recognition, {CVPR} 2025, Nashville, TN, USA, June 11-15, 2025}.\hskip 1em plus 0.5em minus 0.4em\relax Computer Vision Foundation / {IEEE}, 2025, pp. 17\,394--17\,404.

\bibitem{yang2025gripperkeyposeobjectpointflow}
\BIBentryALTinterwordspacing
Y.~Yang, Z.~Cai, Y.~Tian, J.~Zeng, and J.~Pang, ``Gripper keypose and object pointflow as interfaces for bimanual robotic manipulation,'' \emph{CoRR}, vol. abs/2504.17784, 2025. [Online]. Available: \url{https://doi.org/10.48550/arXiv.2504.17784}
\BIBentrySTDinterwordspacing

\bibitem{jia2024lift3dfoundationpolicylifting}
Y.~Jia, J.~Liu, S.~Chen, C.~Gu, Z.~Wang, L.~Luo, X.~Li, P.~Wang, Z.~Wang, R.~Zhang, and S.~Zhang, ``Lift3d policy: Lifting 2d foundation models for robust 3d robotic manipulation,'' in \emph{Proceedings of the IEEE/CVF Conference on Computer Vision and Pattern Recognition (CVPR)}, 2025, pp. 17\,347--17\,358.

\bibitem{chen2024sugarpretraining3dvisual}
\BIBentryALTinterwordspacing
S.~Chen, R.~Garcia, I.~Laptev, and C.~Schmid, ``Sugar: Pre-training 3d visual representations for robotics,'' in \emph{Proceedings of the IEEE/CVF Conference on Computer Vision and Pattern Recognition (CVPR)}, 2024, pp. 18\,049--18\,060. [Online]. Available: \url{https://arxiv.org/abs/2404.01491}
\BIBentrySTDinterwordspacing

\bibitem{kerr2024robotrobotdoimitating}
J.~Kerr, C.~M. Kim, M.~Wu, B.~Yi, Q.~Wang, K.~Goldberg, and A.~Kanazawa, ``Robot see robot do: Imitating articulated object manipulation with monocular 4d reconstruction,'' 2024.

\bibitem{jaegle2022perceiverio}
\BIBentryALTinterwordspacing
A.~Jaegle, S.~Borgeaud, J.-B. Alayrac, C.~Doersch, C.~Ionescu, D.~Ding, S.~Koppula, D.~Zoran, A.~Brock, E.~Shelhamer, O.~J. Henaff, M.~Botvinick, A.~Zisserman, O.~Vinyals, and J.~Carreira, ``Perceiver {IO}: A general architecture for structured inputs \& outputs,'' in \emph{The Tenth International Conference on Learning Representations (ICLR)}, 2022. [Online]. Available: \url{https://arxiv.org/abs/2107.14795}
\BIBentrySTDinterwordspacing

\bibitem{darcet2024visiontransformersneedregisters}
\BIBentryALTinterwordspacing
T.~Darcet, M.~Oquab, J.~Mairal, and P.~Bojanowski, ``Vision transformers need registers,'' in \emph{The Twelfth International Conference on Learning Representations}, 2024. [Online]. Available: \url{https://arxiv.org/abs/2309.16588}
\BIBentrySTDinterwordspacing

\bibitem{sucan2012the-open-motion-planning-library}
I.~A. {\c{S}}ucan, M.~Moll, and L.~E. Kavraki, ``The {O}pen {M}otion {P}lanning {L}ibrary,'' \emph{{IEEE} Robotics \& Automation Magazine}, vol.~19, no.~4, pp. 72--82, 2012, \url{https://ompl.kavrakilab.org}.

\bibitem{chi2024diffusionpolicyvisuomotorpolicy}
C.~Chi, Z.~Xu, S.~Feng, E.~Cousineau, Y.~Du, B.~Burchfiel, R.~Tedrake, and S.~Song, ``Diffusion policy: Visuomotor policy learning via action diffusion,'' \emph{The International Journal of Robotics Research}, 2024.

\bibitem{zhao2023learningfinegrainedbimanualmanipulation}
\BIBentryALTinterwordspacing
T.~Z. Zhao, V.~Kumar, S.~Levine, and C.~Finn, ``Learning fine-grained bimanual manipulation with low-cost hardware,'' in \emph{Robotics: Science and Systems XIX, Daegu, Republic of Korea, July 10-14, 2023}, K.~E. Bekris, K.~Hauser, S.~L. Herbert, and J.~Yu, Eds., 2023. [Online]. Available: \url{https://doi.org/10.15607/RSS.2023.XIX.016}
\BIBentrySTDinterwordspacing

\bibitem{huang2023voxposercomposable3dvalue}
\BIBentryALTinterwordspacing
W.~Huang, C.~Wang, R.~Zhang, Y.~Li, J.~Wu, and L.~Fei-Fei, ``Voxposer: Composable 3d value maps for robotic manipulation with language models,'' in \emph{Proceedings of The 7th Conference on Robot Learning}, ser. Proceedings of Machine Learning Research, J.~Tan, M.~Toussaint, and K.~Darvish, Eds., vol. 229.\hskip 1em plus 0.5em minus 0.4em\relax PMLR, 2023, pp. 540--562. [Online]. Available: \url{https://proceedings.mlr.press/v229/huang23b.html}
\BIBentrySTDinterwordspacing

\end{thebibliography}

\end{document}